\documentclass[journal,transmag]{IEEEtran}
 
\usepackage{amsmath,epsfig}
\usepackage{amsmath,amssymb,latexsym,xspace,float,multirow,fancyvrb,xr-hyper,xr}
\usepackage[breaklinks=true]{hyperref}
\usepackage{amsmath,latexsym,epsfig}
\usepackage{theorem}
\usepackage{paralist}
\usepackage{afterpage}
\usepackage{epstopdf}
\usepackage{setspace}
\usepackage{enumerate}
\usepackage{array}
\usepackage{float}
\usepackage{algorithm}
\usepackage{algorithmic}
\usepackage{setspace}
\usepackage{enumerate}
\usepackage{multirow}
\usepackage{cite}
\usepackage{enumitem}
\usepackage{mathrsfs}
\usepackage{amsfonts}
\usepackage{amssymb}
\usepackage{subcaption} 
\usepackage{graphicx, stfloats}
\usepackage{amstext}

\floatstyle{ruled}
\newfloat{Algorithm}{tb}{lox}
\floatname{Algorithm}{Algorithm}
\newcommand*{\LargerCdot}{\raisebox{-0.25ex}{\scalebox{1.5}{$\cdot$}}}
\newtheorem{theorem}{Theorem} 
\newtheorem{lemma}[theorem]{Lemma}

\ifCLASSINFOpdf
\else
\fi
\begin{document}
%
\title{Kernel Selection using Multiple Kernel Learning and Domain Adaptation in Reproducing Kernel Hilbert Space, for Face Recognition under Surveillance Scenario}


\author{\IEEEauthorblockN{Samik Banerjee\IEEEauthorrefmark{1} and
Sukhendu Das\IEEEauthorrefmark{1}}\\
\IEEEauthorblockA{\IEEEauthorrefmark{1}Department of Computer Science and Engineering,
Indian Institute of Technology, Madras, Chennai 600036, INDIA}
}

%



\IEEEtitleabstractindextext{%
\begin{abstract}
Face Recognition (FR) has been the interest to several researchers over the past few decades due to its passive nature of biometric authentication. Despite high accuracy achieved by face recognition algorithms under controlled conditions, achieving the same performance for face images obtained in surveillance scenarios, is a major hurdle. Some attempts have been made to super-resolve the low-resolution face images  and improve the contrast, without considerable degree of success. The proposed technique in this paper tries to cope with the very low resolution and low contrast face images obtained from surveillance cameras, for FR under surveillance conditions. For Support Vector Machine classification, the selection  of appropriate kernel has been a widely discussed issue in the research community. In this paper, we propose a novel kernel selection technique termed as MFKL (Multi-Feature Kernel Learning) to obtain the best feature-kernel pairing. Our proposed technique employs a effective kernel selection by Multiple Kernel Learning (MKL) method, to choose the optimal kernel to be used along with unsupervised domain adaptation method in the Reproducing Kernel Hilbert Space (RKHS), for a solution to the problem. Rigorous experimentation has been performed on three real-world surveillance face datasets : FR\_SURV \cite{rudrani2011face}, SCface \cite{grgic2011scface} and ChokePoint \cite{wong_cvprw_2011}. Results have been shown using Rank-1 Recognition Accuracy, ROC and CMC measures. Our proposed method outperforms all other recent state-of-the-art techniques by a considerable margin.
\end{abstract}

}

\maketitle

\IEEEdisplaynontitleabstractindextext

%
\IEEEpeerreviewmaketitle

\section{Introduction}
Face Recognition (FR) is a classical problem which is far from being solved. Face Recognition has a clear advantage of being natural and passive over other biometric techniques requiring co-operative subjects. Most face recognition algorithms perform well under a controlled environment. A face recognition system trained at a certain resolution, illumination and pose, recognizes faces under similar conditions with very high accuracy. In contrary, if the face of the same subject is presented with considerable change in environmental conditions, then such a face recognition system fails to achieve a desired level of accuracy. So, we aim to find a solution to the face recognition under unconstrained environment.

Face images obtained by an outdoor panoramic surveillance camera, are often confronted with severe degradations (e.g., low-resolution, blur, low-contrast, interlacing and noise). This significantly limits the performance of face recognition systems used for binding ``security with surveillance'' applications. Here, images used for training are usually available beforehand which are taken under a well controlled environment in an indoor setup (laboratory, control room), whereas the images used for testing are captured when a subject comes under a surveillance scene. With ever increasing demands to combine ``security with surveillance'' in an integrated and automated framework, it is necessary to analyze samples of face images of subjects acquired by a surveillance camera from a long distance. Hence the subject must be accurately recognized from a low resolution, blurred and degraded (low contrast, aliasing, noise) face image, as obtained from the surveillance camera. These face images are difficult to match because they are often captured under non-ideal conditions. Thus, face recognition in surveillance scenario is an important and emerging research area which motivates the work presented in this paper.  

Performance of most classifiers degrade when both the resolution and contrast of face templates used for recognition are low. There have been many advancement in this area during the past decade, where attempts have been made to deal with this problem under an unconstrained environment.  For surveillance applications, a face recognition system must recognize a face in an unconstrained environment without the notice of the subject. Degradation of faces is quite evident in the surveillance scenario due to low-resolution and camera-blur. Variations in the illuminating conditions of the faces not only reduces the recognition accuracy but occasionally degrades the performance of face detection which is the first step of face recognition. The work presented in this paper deals with such issues involved in FR under surveillance conditions.

In the work presented in this paper, the face samples from both gallery and probe are initially passed through a robust face detector, the \textit{Chehra} face tracker, to find a tightly cropped face image. A domain adaptation (DA) based algorithm, formulated using eigen-domain transformation is designed to bridge the gap between the features obtained from the gallery and the probe samples. A novel Multiple kernel Learning (MKL) based learning method, termed MFKL (Multi-Feature Kernel Learning), is then used to obtain an optimal combination (pairing) of the feature and the kernel for FR. The novelty of the work presented in this paper is the optimal pairing of feature and kernel to provide best performance with DA based learning for FR. Results of performance analysis on three real-world surveillance datasets (SCFace \cite{grgic2011scface}, FR\_SURV \cite{rudrani2011face}, ChokePoint \cite{wong_cvprw_2011}) exhibit the superiority of our proposed method of kernel selection by MFKL, with DA in Reproducing Kernel Hilbert Space (RKHS) \cite{dym1989j}.

\section{Discussion on Related Work}
\label{rw}
The problem of automatic face recognition consists of three key steps/subtasks: (1) detection and rough normalization of faces, (2) feature extraction and accurate normalization of faces, (3) identification and/or verification. These different subtasks are not totally isolated. For example, the discriminating facial features (eyes, nose, mouth) used for face recognition are often used in face detection. Face detection and feature extraction can be achieved simultaneously. Recent advances of face detection, MKL and DA are discussed henceforth.

The most widely used algorithm for FD (face detection) was proposed by Viola et al. in \cite{viola2004robust}. The algorithm proposed in \cite{viola2004robust} is based on a simple and efficient classifier which is built using the AdaBoost learning algorithm to select a small number of critical visual features from a very large set of potential features. The authors proposed a method for combining classifiers in a “cascade” which allows background regions of the image to be quickly discarded while spending more computation on promising face-like regions. A very recent state-of-the-art technique proposed by Zhu et al. \cite{zhu2012face} presents a unified model for face detection, pose estimation, and landmark estimation in real-world, cluttered images. Their proposed model is based on a mixture of trees with a shared pool of parts which model every facial landmark as a part and use global mixtures to capture topological changes due to viewpoint. The tree-structured models are surprisingly effective at capturing global elastic deformation, while being easy to optimize unlike dense graph structures. Yow et al. \cite{yow1997feature} also proposed a feature-based algorithm for detecting faces that is sufficiently generic and is also easily extensible to cope with more demanding variations of the imaging conditions. The algorithm detects feature points from the image using spatial filters and groups them into face candidates using geometric and gray level constraints. A probabilistic framework is then used to reinforce probabilities and to evaluate the likelihood of the candidate as a face. Our proposed method uses a set of 49 fiducial landmark points detected by the Chehra face detector \cite{asthanaincremental}.  The incremental training of discriminative models used by Chehra is not only important for building person-specific models but also to update a generic model in case a new annotated data arrives, since the training procedure is very expensive and time consuming. In particular, incremental training of discriminative models use a cascade of linear regressors to learn the mapping from facial texture to the shape.

While classical kernel-based classifiers are based on a single kernel, in practice it is often desirable to form classifiers based on combinations of multiple kernels. Bach et al. \cite{bach2004multiple} proposed the sequential minimal optimization (SMO) techniques that are essential in large-scale implementations of the SVM for large number of kernels. Shrivastava et al. \cite{shrivastava2014multiple} proposed a multiple kernel learning (MKL) algorithm that is based on the sparse representation-based classification (SRC) method efficiently representing the nonlinearities in the high-dimensional feature space, based on the kernel alignment criteria. This method uses a two step training method to learn the kernel weights and sparse codes. At each iteration, the sparse codes are updated first while fixing the kernel mixing coefficients, and then the kernel mixing coefficients are updated while fixing the sparse codes. These two steps are repeated until a stopping criteria is met. Lanckriet et al. \cite{lanckriet2004statistical} considered conic combinations of kernel matrices for classification, leading to a convex quadratically constrained quadratic program. Sonnenburg et al. \cite{sonnenburg2006large} show that it can be rewritten as a semi-infinite linear program that can be efficiently solved by recycling the standard SVM implementations. Moreover, Sonnenburg et al. \cite{sonnenburg2006large} generalize the formulation and method to a larger class of problems, including regression and one-class classification. The method proposed in our paper uses Structured MKL \cite{xu2010simple}, which is a modified version of simpleMKL\cite{alain2008simple}, through a primal formulation involving a weighted L2-norm regularization. 

Domain adaptation has gained enormous importance in the recent past. One of the popular solutions of this problem is to weigh each instance in the source domain appropriately such that, when the weighted instances of the source domain are used for training the expected loss of classifiers tested for target domain is minimized. Some of the works where instance weights of source domain have been calculated are \cite{sugiyama2008direct}, \cite{dai2007transferring}, \cite{jiang2007instance}, \cite{pathak2009learning}. Sugiyama et al. \cite{sugiyama2008direct} proposed a method to calculate weights for instances in source domain, by using a convex optimization framework which minimizes the KL-divergence between the source and the target domain. There have been some work for building robust classifiers for domain adaptation \cite{aytar2011tabula}, \cite{yang2007cross}, \cite{duan2012domain}. Yang et. al. \cite{yang2007cross} has proposed a method to calculate weights for each instances in source domain, which are used to effectively retrain a pre-learned SVM for target domain. Domain Adaptive Machine (DAM), proposed by Duan et al. \cite{duan2012domain}, learns a robust decision function for labeling the instances in target domain, by leveraging a set of base classifies learned on multiple source domains. Pan et al. \cite{pan2011domain} proposed transfer component analysis (TCA), which minimizes the disparity of distribution by considering the difference of means between two domains and it also preserves local geometry of underlying manifold. The study of subspaces spanned by the input data have found important applications in computer vision tasks \cite{turk1991eigenfaces}. These subspaces identify important properties of the data which can be used to model the data. Subspace based modeling have also been widely used for the task of visual domain adaptation \cite{fernando2013unsupervised}, \cite{baktashmotlagh2013unsupervised}. Fernando et. al. \cite{fernando2013unsupervised} has calculated a subspace using eigen-vectors of two domains such that the basis vectors of transformed source and target domains are aligned. Manifold alignment has also been used for domain adaptation earlier. Wang et al. \cite{wang2011heterogeneous} has considered the manifold of each domain and estimated a latent space, where the manifolds of both the domains are similar to each other. However, the structures or the distributions of the domains have not been considered in this case. In this paper we transform the source domain data into the target domain based on the eigen vectors of both the domains. Samanta et al. in \cite{samanta2014modeling} describe a new method of unsupervised domain adaptation (DA) using the properties of the sub-spaces spanning the source and target domains, when projected along a path in the Grassmann manifold.  A new technique of unsupervised domain adaptation based on eigenanalysis in kernel space, for the purpose of categorisation tasks, had also been proposed in \cite{samanta2015unsupervised}. Here, the authors propose a transformation of data in source domain, such that the eigenvectors and eigenvalues of the transformed source domain become similar to that of the target domain. They extend this idea to the RKHS\cite{dym1989j}, which enables to deal with non-linear transformation of source domain. To handle non-linearity of the data, we also rely on the DA projected in RKHS.

Recently, face recognition research in real-life surveillance has become very popular. For high data transmission speed and easy data storage, surveillance cameras generally produce images in low resolution, and face images captured directly by surveillance cameras are usually very small. Besides, images taken by surveillance cameras are generally with noises and corruptions, due to the uncontrolled circumstances and distances. Zou et al. \cite{zou2012very} proposed a super-resolution approach to increase the recognition performance for very low-resolution face images. They employ a minimum mean square error estimator to learn the relationship between low and high resolution training pairs. A further discriminative constraint is added to the learning approach using the class label information. Biswas et al.  \cite{biswas2012multidimensional} proposed a matching algorithm through using Multidimensional Scaling (MDS). In their approach both the low and high resolution training pairs are projected into a kernel space. Transformation relationship is then learned in the kernel space through iterative majorization algorithm, which is used to match the low-resolution test faces to the high-resolution gallery faces. Similarly, Ren et al. \cite{ren2012coupled} proposed the Coupled Kernel Embedding approach, where they map the low and high resolution face images onto different kernel spaces and then transform them to a learned subspace for recognition.

Rudrani et al. in \cite{rudrani2011face} proposed an approach with the combination of partial restoration (using super-resolution) of probe samples and degradation of gallery samples. The authors also proposed a outdoor surveillance dataset, FR\_SURV \cite{rudrani2011face} for evaluating their approach. In our previous work proposed in \cite{banerjee2014face}, we aim to bridge the gap of resolution and contrast using super-resolution and contrast stretching on the probe samples and degrading the gallery samples by downsampling the gallery samples and introducing a Gaussian blur to the downsampled images. In addition to these measures, we had also proposed a DA technique based on an eigen-domain transformation to make the distributions of the features obtained from the gallery and probe samples identical. 

In the following sections, we first briefly present a few technical background details, followed by our proposed framework and then our experimental results of the proposed technique.

\section{Brief Technical Background}
\subsection{Multiple kernel learning problem}
\label{mkls}
A seemingly unrelated problem in machine learning, the problem of \emph{multiple kernel learning} is in fact intimately connected with sparsity-inducing norms by duality. It actually corresponds to the most natural extension of sparsity to Reproducing Kernel Hilbert Spaces \cite{dym1989j}. It can be shown that for a large class of norms and, among them, many sparsity-inducing norms, there exists for each of them a corresponding multiple kernel learning scheme, and, vice-versa, each multiple kernel learning scheme defines a new norm.

In the multiple kernel learning problem, $n$ data points $(x_i,y_i)$ are given, where $x_i \in \mathcal{X}$ for some input space $\mathcal{X} \in \mathbf{R}^l$, and where $y_i \in \{-1, 1\}$. Also, assume that $m$ kernels are given $K_j \in \mathbf{R}^{n \times n}$, which are assumed to be symmetric positive semidefinite, obtained from evaluating a kernel function on the data ${x_i}$. Consider the problem of learning the best linear combination $\sum_{j=1}^{m} \eta_j K_j$ of kernels $K_j$ with non-negative coefficients $\eta_j > 0$ and with a trace constraint $\text{tr} \sum_{j=1}^{m} \eta_j K_j = \sum_{j=1}^{m} \eta_j \text{tr} K_j = c$, where $c > 0$ is fixed. Lanckriet et al. \cite{lanckriet2004learning} show that this setup yields the following optimization problem:
\begin{equation}
\label{eq:mkl}
\begin{split}
\text{min} \hspace{4pt} & \zeta - 2 e^T\alpha \\
\text{w.r.t.} \hspace{4pt} & \zeta \in \mathbf{R}, \alpha \in \mathbf{R}^n \\
\text{s.t.} \hspace{4pt} & 0 \leq \alpha \leq C, \alpha^T y=0, \text{and} \\
& \alpha^T D(y) K_j D(y) \alpha \leq \frac{\text{tr} K_j}{c} \zeta , j \in \{1,...,m\}
\end{split}
\end{equation}

where $D(y)$ is a diagonal matrix with all diagonal elements as $y, e \in \mathbf{R}^n$ the vector of all ones, and $C$ a positive constant. The coefficients $\eta_j$ are recovered as Lagrange multipliers for the constraints $\alpha^T D(y) K_j D(y) \alpha \le \frac{\text{tr} K_j}{c} \zeta$. 

\subsection{Support Kernel Learning Machine}
\label{sec:mfkl}

Consider a decomposition of $\mathbf{R}^l$ into $p$ blocks (subspaces): $\mathbf{R}^l = \mathbf{R}^{l_1} \times ... \times \mathbf{R}^{l_p}$, such that, $x_i^f \in \mathbf{R}^{l_f}$, ($x_i^f$ is a vector). The classification algorithm, "support kernel machine", as proposed by Bach et al. \cite{bach2004multiple}, is exactly the dual of the problem defined in equation \ref{eq:mkl}.

The aim is to obtain a linear classifier of the form $y = \text{sign}(w^Tx+b)$, where $w$ has identical block-level decomposition as $w = (w_1, ... , w_p) \in \mathbf{R}^{l_1 + ... + l_p}$. To induce sparsity in the vector $w$ at the level of blocks, such that most of its components $w_i$ go to zero, the $l1$-norm of the square of a weighted block of $w$, $(\sum_{f=1}^p d_f \|w_f \|_2)^2$, is minimized, where the $l2$-norm is used within every block. The primal problem  is thus:

\begin{equation}
\label{eq:mfkl_p}
\begin{split}
\text{min} \hspace{4pt} & \frac{1}{2} (\sum_{f=1}^p d_f \|w_f \|_2)^2 + C \sum_{i=1}^n \xi_i \\
\text{w.r.t.} \hspace{4pt} & w = (w_1, ... , w_p) \in \mathbf{R}^{l_1 + ... + l_p}, b \in \mathbf{R}, \xi \in \mathbf{R}^n_{+} \\
\text{s.t.} \hspace{4pt} & y_i(\sum_f w^T_f x_i^f + b) \geq 1 - \xi_i, \forall i \in \{1,...,n\}.
\end{split}
\end{equation}

The problem posed in equation \ref{eq:mfkl_p} is treated as a second-order cone program (SOCP) \cite{lobo1998applications}, which yields the following dual:

\begin{equation}
\label{eq:mkl_d}
\begin{split}
\text{min} \hspace{4pt} & \frac{1}{2} \gamma^2 - \alpha^T e \\
\text{w.r.t.} \hspace{4pt} & \gamma \in \mathbf{R}, \alpha \in \mathbf{R}^n \\
\text{s.t.} \hspace{4pt} & 0 \leq \alpha \leq C, \alpha^T y=0 \text{, and}\\
& \| \sum_{i} \alpha_i y_i x_i^f \|_2 \leq d_f \gamma, \forall f \in \{1,...,p\}.
\end{split}
\end{equation}

To induce sparsity in the formulation, the following KKT conditions play an instrumental role to give the complementary slackness equations:

\begin{enumerate}
\item $\alpha_i(y_i(\sum_f w^T_f x_i^f + b)-1+ \zeta_i) = 0, \forall i$
\item $(C-\alpha_i)\zeta_i = 0, \forall i$
\item ${w_f \choose {\|w_f\|_2}}^T {{- \sum_i \alpha_i y_i x_i^f} \choose {d_f \gamma}}=0, \forall f$
\item $\gamma (\sum d_f t_f -\gamma) = 0$
\end{enumerate}

In RKHS, the data points $x_i$ are embedded in an Euclidean space via a mapping $\phi : \mathcal{X} \rightarrow \mathbf{R}^c$ and $\phi(x)$ has $m$ distinct block components $\phi(x) = (\phi_1(x), ... , \phi_m(x))$. Following the usual recipe for kernel methods, this embedding is performed implicitly, by specifying the inner product in $\mathbf{R}^c$ using a kernel function, which in this case is the sum of individual kernel functions on each block component:

\begin{equation}
\label{eq:mfkl_k}
\begin{split}
k(x_i,x_j) = \phi(x_i)^T \phi(x_j)  &= \sum_{s=1}^{m} \phi_s(x_i)^T \phi_s(x_j) \\
&= \sum_{s=1}^{m} k_s(x_i,x_j)
\end{split}
\end{equation}


The minimization task described in equation \ref{eq:mfkl_p} is kernelized \cite{dym1989j} using this kernel function (equation \ref{eq:mfkl_k}). In particular, considering the dual of the problem in equation \ref{eq:mfkl_p} and substituting the kernel function for the inner products in equation \ref{eq:mkl_d}, one obtains:

\begin{equation}
\label{eq:mkl_dk}
\begin{split}
\text{min} \hspace{4pt} & \frac{1}{2} \gamma^2 - \alpha^T e \\
\text{w.r.t.} \hspace{4pt} & \gamma \in \mathbf{R}, \alpha \in \mathbf{R}^n \\
\text{s.t.} \hspace{4pt} & 0 \leq \alpha \leq C, \alpha^T y=0 \\
& (\alpha^T D(y) K_j(y) \alpha) ^ {\frac{1}{2}}  \leq d_j \gamma, \forall j.
\end{split}
\end{equation}

where $K_j$ is the $j$-th Gram matrix of the points $\{x_i\}$ corresponding to the $j$-th kernel.

Equation \ref{eq:mkl_dk} is rearranged to yield an equivalent formulation in which the quadratic constraints are incorporated into the objective function:

\begin{equation}
\label{eq:mkl_s}
\begin{split}
\text{min} \hspace{4pt} & \text{max}_j \frac{1}{2d^2_j} \alpha^T D(y) K_j D(y) \alpha - \alpha^T e \\
\text{w.r.t.} \hspace{4pt} & \alpha \in \mathbf{R}^n \\
\text{s.t.} \hspace{4pt} & 0 \leq \alpha \leq C, \alpha^T y=0 \\
\end{split}
\end{equation}

Let, $J_j(\alpha)$ denote $\frac{1}{2d^2_j} \alpha^T D(y) K_j D(y) \alpha - \alpha^T e$ and $J(\alpha) = \text{max}_j J_j(\alpha)$. Minimization of $J(\alpha)$ now reduces to an convex optimization problem as described by equation \ref{eq:mkl_s}, as $J(\alpha)$ is a non-differentiable convex function subject to linear constraints.

\subsection{Domain Adaptation (DA) based on Eigen Domain transformation \cite{banerjee2014face}}
\label{da}
Given a data, its distribution can be estimated using the covariance matrix or Eigen-vectors. Hence, if the Eigen-vectors of two datasets are same, then we can say that the distributions of the two datasets are approximately similar to each other. Hence, in this paper we aim to transform the source domain in such a way that the Eigen-vectors of the transformed source domain is same as that of the target domain. We extend this idea of transformation of source domain in Reproducing Kernel Hilbert Space (RKHS) \cite{dym1989j} to handle non-linear transformation of data, when necessary. In the following sub-sections, we give the mathematical details necessary for the unsupervised method of domain adaptation (as in our earlier work in \cite{banerjee2014face}).

\subsubsection{Learning the Transformation}
\label{lda}
Let $S$ and $T$ be the source and target domains having $n_S$ and $n_T$ number of instances respectively and $\tilde{S}$ be the transformed source domain. Let the $i^{th}$ and $j^{th}$ instance of $S$ and $T$ be represented as $S_i$ and $T_j$ respectively. Let, the principal components of $S$ and $T$ be denoted by the matrices $U_S$ and $U_T$ respectively, where the $i^{th}$ column represents the Eigen-vector corresponding to the $i^{th}$ largest Eigen-value. Then the principal components of $T$ \& $\tilde{S}=SU_SU_T^T$ are the same, which is shown in Lemma \ref{lemma1}. Hence, a simple transformation of the source domain can be obtained by multiplying it with the Eigen-vectors of the source and target domains. 

\begin{lemma}
  \label{lemma1}
  If $U_P$ and $U_Q$ are the principal components of two datasets $P$ and $Q$ respectively, then the principal component of $QU_QU_P^T$ is $U_P$.
  \end{lemma}
  Let $\Lambda_P$ and $\Lambda_Q$ be two diagonal matrices whose diagonal elements are the eigen-values of $P$ and $Q$ respectively. The covariance matrices of $P$ and $Q$ can be written as $U_P\Lambda_PU_P^T$ and $U_Q\Lambda_QU_Q^T$ respectively. Now, the covariance matrix of $\hat{Q}=(QU_QU_P^T)$ is given as:
  \begin{eqnarray*}
  \hat{Q}^T\hat{Q}&=&U_PU_Q^TQ^TQU_QU_P^T \\
  &=&U_PU_Q^TU_Q\Lambda_QU_Q^TU_QU_P^T=U_P\Lambda_QU_P^T
  \end{eqnarray*}
  This shows that the principle components of $QU_QU_P^T$ is $U_P$.

This can be followed by shifting of $\tilde{S}$ such that its mean is same as that of the target domain data.

\subsubsection{Extension to the RKHS}

The above formulation of DA performs linear transformation of the source domain data. In order to handle non-linear transformation of data, we extend the formulation to RKHS. If $\Phi(.)$ is a universal kernel function, then in kernel space the source and target domains are $\Phi({S})$ and $\Phi({T})$ respectively. Let $K_{SS}$ and $K_{TT}$ be the Gram matrices of $\Phi(S)$ and $\Phi(T)$ respectively ($K_{SS}=\Phi(S)\Phi(S)^T$, $K_{ST}=\Phi(S) \Phi(T)^T$ and $K_{TT}=\Phi(T) \Phi(T)^T$). Let, $U_S^{\Phi}$ and $U_T^{\Phi}$ be the principle components of $\Phi(S)$ and $\Phi(T)$ respectively. Also, let $V_S^{\Phi}$ and $V_T^{\Phi}$ be the eigen-vectors of $K_{SS}$ and $K_{TT}$ respectively. Then we can write,
 \begin{eqnarray}
    U^{\Phi}_S &=& \Phi(S)^TV_S^{\Phi} \\
    U^{\Phi}_T &=& \Phi(T)^TV_T^{\Phi}
 \end{eqnarray}
 Then the transformed source domain in RKHS is given by:
 \begin{equation}
 \Phi(\tilde{S})=\Phi(S)U^{\Phi}_S{U_T^{\Phi}}^T=K_{SS}V_S^\Phi V_T^{\Phi T}\Phi(T)
 \end{equation}
 The corresponding Gram matrices are given by:
 \begin{equation}
K_{\tilde{S}\tilde{S}} = \Phi(\tilde{S})\Phi(\tilde{S})^T =K_{SS}V_S^\Phi V_T^{\Phi T}K_{TT}V_T^\Phi V_S^{\Phi T}K_{SS} 
\end{equation}
\begin{equation}
K_{\tilde{S}T}=\Phi(\tilde{S})\Phi(T)^T = K_{SS}V_S^\Phi V_T^{\Phi T}K_{TT}
 \end{equation}
 Once we obtain the Gram matrices, we need to modify them appropriately such that the mean of the transformed source domain is same as that of the target domain. If $o_S \in \mathbb{R}^{n_S}$ and $o_T \in \mathbb{R}^{n_T}$ be two vectors with all elements as $1/{n_S}$ and $1/{n_T}$ respectively, then the mean of transformed source domain and target domain in RKHS is given by $o_S^T\Phi(\tilde{S})$ and $o_T^T\Phi(T)$ respectively. Let the $i^{th}$ row of a gram matrix $K$ is denoted by $K(i,\LargerCdot)$ and let the $j^{th}$ column of $K$ is denoted by $K_(\LargerCdot, j)$. Let $K(i,j)$ denote the value corresponding to the $i^{th}$ row and $j^{th}$ column of $K$. 
 
If $\hat{K}_{\tilde{S}\tilde{S}}$ represents the mean shifted Gram matrix, then each element of this matrix can be calculated as:
 \begin{eqnarray}
 \hat{K}_{\tilde{S}\tilde{S}}(i,j) &=& (\Phi(\tilde{S}_i)-o^T_S\Phi(\tilde{S})+o^T_T\Phi(T))\times \nonumber \\ 
&& \hspace{20mm}(\Phi(\tilde{S}_i)-o^T_S\Phi(\tilde{S})+o^T_T\Phi(T))^T \nonumber \\
&=& K_{\tilde{S}\tilde{S}}(i,j)- K_{\tilde{S}\tilde{S}}(i,\LargerCdot)o_S +K_{\tilde{S}T}(i,\LargerCdot)o_T \nonumber \\ 
&&  -o_S^TK_{\tilde{S}\tilde{S}}(\LargerCdot,j) +o^T_SK_{\tilde{S}\tilde{S}}o_S-o^T_SK_{\tilde{S}T}o_T \nonumber \\
&& +o^T_TK_{T\tilde{S}}(\LargerCdot,j)-o^T_TK_{T\tilde{S}}o_S + o^T_TK_{TT}o_T
 \end{eqnarray}
 
 Similarly, each element of the mean-shifted Gram matrix $\hat{K}_{\tilde{S}T}$ can be calculated as:
 \begin{eqnarray}
  \hspace{-3mm} \hat{K}_{\tilde{S}T}(i,j) \hspace{-3mm}&=& \hspace{-2mm} (\Phi(\tilde{S}_i)-o^T_S\Phi(\tilde{S})+o^T_T\Phi(T)) \times \Phi(T_j)^T \nonumber \\
\hspace{-3mm} &=& \hspace{-3mm} K_{\tilde{S}T}(i,j)- o^T_S K_{\tilde{S}T}(\LargerCdot,j) + o^T_TK_{TT}(\LargerCdot,j)
  \end{eqnarray}

\subsubsection{Classification}

Once we obtain the Gram matrices $\hat{K}_{\tilde{S}\tilde{S}}$ and $\hat{K}_{\tilde{S}T}$, we can calculate the overall Gram matrix
\begin{equation}
\hat{K}=\begin{bmatrix} \hat{K}_{\tilde{S}\tilde{S}} & \hat{K}_{\tilde{S}T} \\ \hat{K}^T_{\tilde{S}T} & K_{TT} \end{bmatrix}
\end{equation}
We can now calculate the Euclidean distance between any two instances ($i$ and $j$) in RKHS, which is given by:
\begin{equation}
dist(i,j)=\hat{K}(i,i)+\hat{K}(j,j)-2\times \hat{K}(i,j)
\end{equation}
Hence, we can now use this distance matrix for classifying test samples using KNN-classifier. The unsupervised method of DA considers the Training set to be the source and the Test set (in a FR dataset) to be the target domains. A subset of a few samples from the target domain are used to estimate the distribution of the target domain (see table \ref{tab:sigda} in section \ref{pfw}). Transformation from source to target is estimated using eigen-analysis of the BOW-based features. The unsupervised method of DA, enhances the performance of FR algorithm on surveillance conditions.

\section{Proposed Framework}
\label{pfw}

The proposed framework as shown in figure \ref{pffig}, has been designed for dealing with the problem of FR under low resolution and low contrast, using multiple kernel learning \cite{bach2004multiple} and DA. The stages are discussed in the following: 

\begin{figure}[!htbp]
\centering
\includegraphics[scale = 0.3]{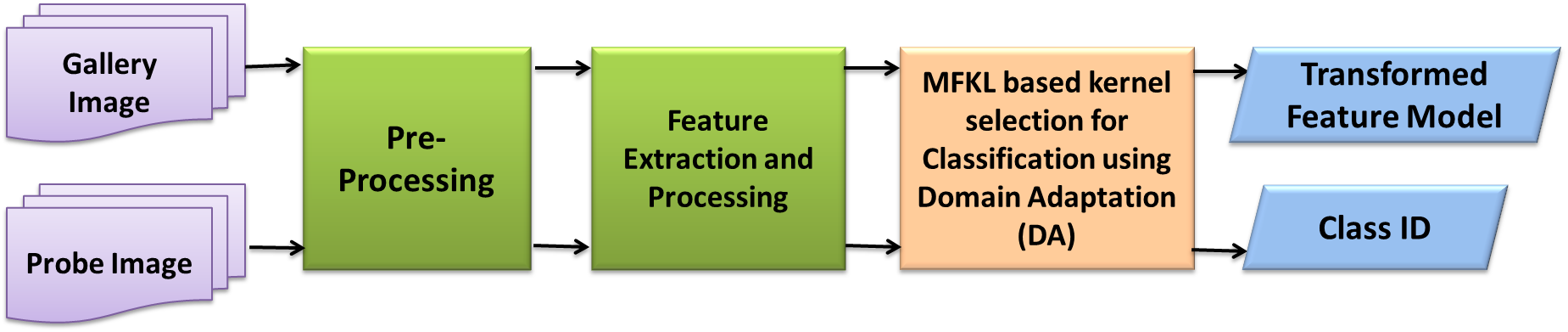}
\caption{The Proposed Framework of FR, designed using DA and MKL.}
\label{pffig}
\end{figure}

\subsection{The pre-processing stage}
\label{pps}

\begin{figure}[!htbp]
\centering
\includegraphics[scale = 0.35]{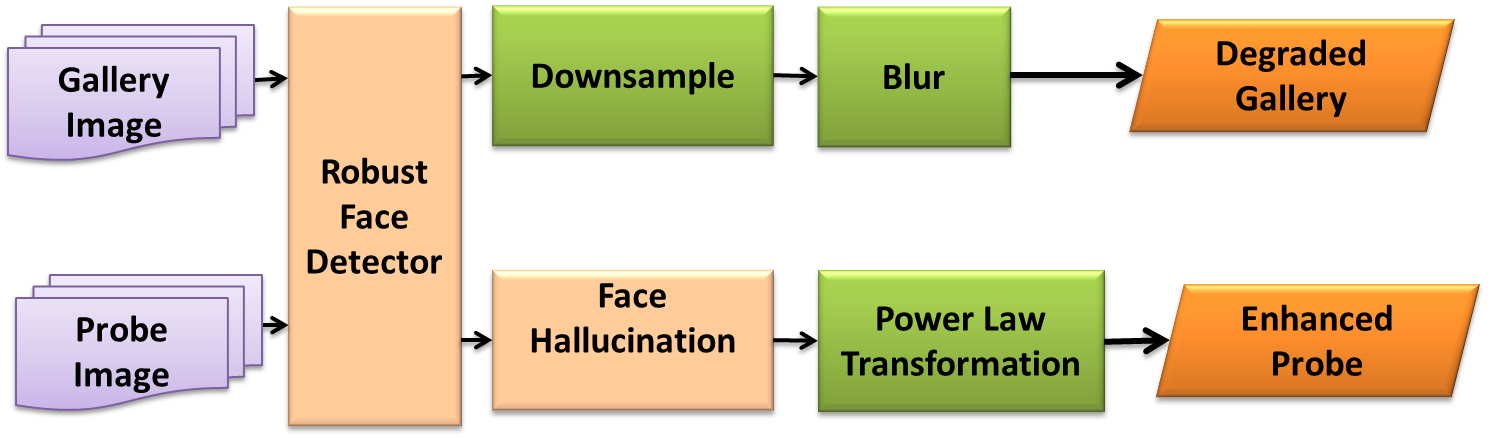}
\caption{The modules of the pre-processing stage in the proposed framework (see figure \ref{pffig}).}
\label{pre_proc}
\end{figure}

Since the faces appear with background and noise, pre-processing of the images is always necessary before they are passed into the feature extraction stage. This treatment of images is called the pre-processing stage as shown in figure \ref{pre_proc} and described below step-wise:

\label{pp}

\subsubsection{Robust Face Detection}
\label{chehra}

The gallery and the probe images are passed through the Chehra face tracker \cite{asthanaincremental} which uses a cascade linear regression for discriminative face alignment. The incremental update of the cascade of linear regression is a very challenging task, since the results from one level have to propagated to the next. Due to this sequential nature of the training procedure, we refer to this method as \textit{Sequential Cascade of Linear Regression} which is learned by a Monte-Carlo procedure \cite{asthanaincremental}. This method deals with the problem of updating a discriminative facial deformable model, a problem that has not been thoroughly studied in the literature. In particular, the strategies to update a discriminative model that is trained by a cascade of regressors is handled by this method. Very efficient strategies to update the model is adopted and it is possible to automatically construct robust discriminative person and imaging condition specific models.

\subsubsection{Face Hallucination}
\label{fh}

The probe samples are upsampled using a state-of-the-art Face hallucination method proposed by Felix et al. in \cite{juefei2015single}. This method of single image face hallucination is based on solo dictionary learning. The core idea of the proposed method \cite{juefei2015single} is to recast the superresolution task as a missing pixel problem, where the low resolution image is considered as its high-resolution counterpart with many pixels missing in a structured manner. A single dictionary is therefore sufficient for recovering the super-resolved image by filling the missing pixels.

\subsubsection{Degradation of Gallery samples}
\label{deg}

\begin{figure} [!htbp]
\centering
\includegraphics[scale = 0.27]{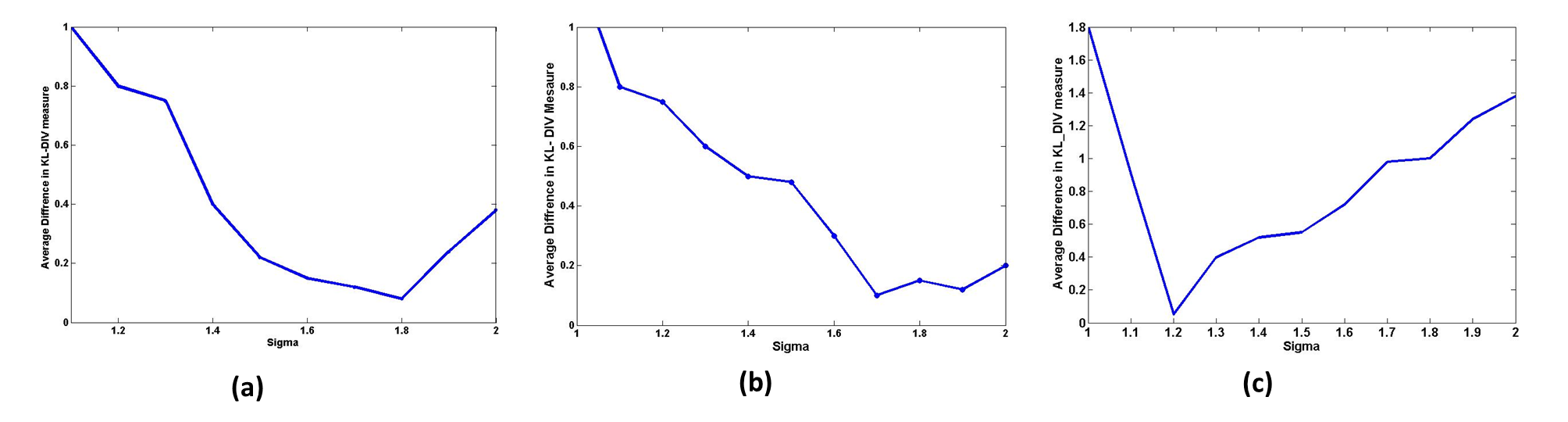}
\caption{Plot of KL-Divergence between degraded gallery and probe images with different values of $\sigma$ used for degradation, in case of (a) FR\_SURV \cite{rudrani2011face}, (b) SCface \cite{grgic2011scface} and (c) ChokePoint \cite{wong_cvprw_2011} datasets. The optimal values of $\sigma$ are obtained as: (a) 1.75, (b) 1.7 and (c) 1.2, respectively for the three datasets.  }
\label{kld}
\end{figure}

The gallery samples are downsampled by two using simple bicubic interpolation method \cite{Gonzalez:92}. The gallery samples are then blurred using an estimated Gaussian blur, $\sigma$. The $\sigma$ is estimated using KL-Divergence\cite{goldberger2003efficient}, between the distributions of the downsampled gallery and the upsampled probe images. The distribution of the target domain is estimated using three samples per class but with no class labels (hence this is unsupervised). The graphs in figure \ref{kld} (a), (b) and (c) show the plots of KL-divergence with increasing values of Gaussian blur kernel, $\sigma$, for the three different datasets, FR\_SURV, SCface and ChokePoint, respectively. The optimal $\sigma_{opt}$ is obtained at 1.75, 1.7 and 1.2 for the three datasets and are recorded in table \ref{tab:sigda}. The gallery samples are degraded using this uniform blur $\sigma_{opt}$ to obtain a degraded gallery.

\begin{table} [!htbp]
\caption{Number of probe samples (per subject) used for estimation of ${\sigma}_{opt}$ and in DA, for three datasets. Class (subject ID) information was made unavailable in both cases (as the method is unsupervised). Only for the estimation of ${\sigma}_{opt}$, the entire gallery is used.}
\begin{center}
\begin{tabular}{|c|c|c|c|}
    \hline
    \multirow{2}{*}{\textbf{Dataset}} & \multicolumn{2}{c|}{\textbf{No. of samples used, for}} & \multirow{2}{*}{\textbf{Values of  ${\sigma}_{opt}$}} \\
    \cline{2-3}
    &\textbf{Estimation of ${\sigma}_{opt}$} & \textbf{DA} & \\
    \hline
    \textbf{SCface} & 5 & \multirow{2}{*}{3} & \multirow{2}{*}{1.7} \\
    \cite{grgic2011scface}& for 30 subjects & &\\
    \hline
   \textbf{FR\_SURV} &  5 & \multirow{2}{*}{3} & \multirow{2}{*}{1.75} \\
    \cite{rudrani2011face}& for 20 subjects &  &\\
    \hline
    \textbf{ChokePoint} &  5 & \multirow{2}{*}{6} & \multirow{2}{*}{1.2}\\
    \cite{wong_cvprw_2011}& for 20 subjects &  &\\
    \hline
\end{tabular}
\end{center}
\label{tab:sigda}
\end{table}

\subsubsection{Power Law Transformation}
\label{plt}

\begin{figure}[!htbp]
\centering
\includegraphics[scale = 0.4]{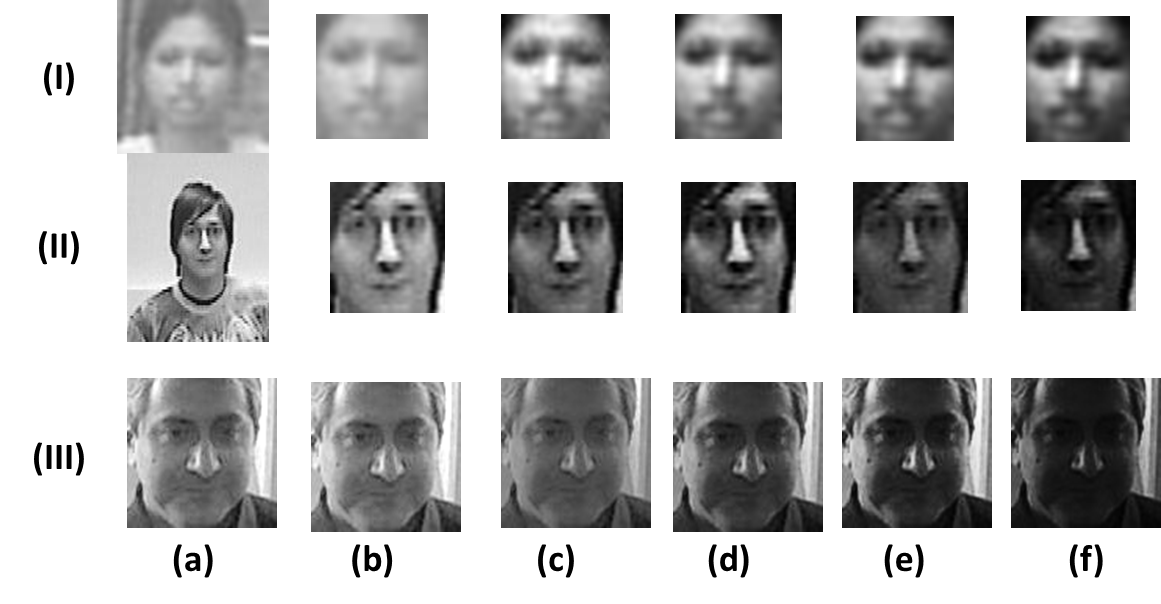}
\caption{(a) Probe images of three subjects; (b) Outputs of CHEHRA \cite{asthanaincremental} process; Gamma-corrected (equation \ref{pl}) images with (c) $\gamma = 1.25$; (d) $\gamma = 1.5$; (e) $\gamma = 1.75$; and (f) $\gamma = 1.9$, from (I) FR\_SURV \cite{rudrani2011face}, (II) SCFace \cite{grgic2011scface} and (III) ChokePoint \cite{wong_cvprw_2011} datasets.}
\label{plfig}
\end{figure}

To cope with the contrast degradation, we perform Power Law transformation \cite{farid2001blind} for contrast stretching the probe images. The transformation function used is:
\begin{equation}
P(i,j) = k. {C(i,j)} ^ \gamma
\label{pl}
\end{equation}
where, $P(i,j)$ and $C(i,j)$ denotes the gray-level pixel values of the input and output image. We use $\gamma = 1.25, k = 1$.
Visual results as shown in figure \ref{plfig} depicts  contrast enhancement of the image for varying values of $\gamma$. We set $\gamma = 1.25$ based on visual observation (empirical).

For probe samples, we use:
\begin{equation}
Probe_{upsampled} = {Probe_{crop}\uparrow v}
\label{ppe3} 
\end{equation}
\begin{equation}
Probe_{transformed}(i,j) = {Probe_{upsampled}(i,j)} ^ \gamma
\label{ppe2} 
\end{equation}
where, $Probe_{crop}$ denotes the cropped probe sample based on Chehra \cite{asthanaincremental}, $\uparrow$ denotes upsampling of the image by a factor $v$ mentioned to the right of it, $Probe_{crop}$ denotes upsampled image and $Probe_{transformed}$ denotes the image that will be used for feature-extraction. The $\gamma$ is the parameter for gamma-correction of the image based on Power Law transformation which is given by equation \ref{pl}.

\subsection{Feature Extraction}
\label{fe}

The feature extraction process in the proposed method is based on the set of features extracted from the degraded gallery and the enhanced probe face images. The set of features used in the methods proposed includes Local Binary Pattern (LBP), Eigen Faces, Fisher Faces, Gabor faces, Weber Faces, Bag of Words (BOW), Vector of Linearly Agregated Descriptors encoding based on Scale Invariant Feature Transform (VLAD-SIFT) \cite{arandjelovic2013all}, Fisher Vector encoding based on SIFT (FV-SIFT) \cite{perronnin2010improving}.

\subsubsection{Eigen Faces}
\label{ef}

This approach \cite{turk1991eigenfaces} of the detection and identification of human faces and describe a working, near-real-time face recognition system which tracks the face of a subject and then recognizes the person by comparing characteristics of the face to those of known individuals. The approach treats face recognition as a two-dimensional recognition problem, taking advantage of the fact that faces are normally upright and thus may be described by a small set of 2-D characteristic views. Face images are projected onto a feature space (“face space”) that best encodes the variation among known face images. The face space is defined by the “eigenfaces”, which are the eigenvectors of the set of faces; they do not necessarily correspond to isolated features such as eyes, ears, and noses.

\subsubsection{Fisher Faces}
\label{ff}

This face recognition algorithm \cite{belhumeur1997eigenfaces} is insensitive to large variation in lighting direction and facial expression. Taking a pattern classification approach, each pixel is considered in an image as a coordinate in a high-dimensional space. The images of a particular face, under varying illumination but fixed pose, lie in a 3D linear subspace of the high dimensional image space if the face is a Lambertian surface without shadowing. However, since faces are not truly Lambertian surfaces and do indeed produce self-shadowing, images will deviate from this linear subspace. Rather than explicitly modeling this deviation, the image is projected into a subspace in a manner which discounts those regions of the face with large deviation. The projection method is based on Fisher’s Linear Discriminant and produces well separated classes in a low-dimensional subspace, even under severe variation in lighting and facial expressions. 

\subsubsection{Gabor Faces}
\label{gf}

The Gabor Feature Classifier (GFC) method \cite{liu2002gabor} employs an enhanced Fisher discrimination model on an augmented Gabor feature vector; which is derived from the Gabor wavelet transformation office images. For the three datasets used for experimentation, table \ref{tab:gab} gives the values of the parameters used, where $v$ represents the different scales used, $n_\mu$ is the number of orientations and $\mu$ is the orientation. Parameter $\sigma$, which determines the ratio of the Gaussian window width to wavelength, is set to $2 \pi$, $k_{max}$ is the wave-vector set to $\pi$ and $f$ the spatial frequency set to $\sqrt{2}$. The Gabor wavelets, whose kernels are similar to the 20 receptive field profiles of the mammalian cortical simple cells, exhibit desirable characteristics of spatial locality and orientation selectivity. As a result, the Gabor transformed face images produce salient local and discriminating features that are suitable for face recognition.

\begin{table}[!htbp]
\caption{Parameters for experimenation in Gabor faces for three datasets. }
\begin{center}
\begin{tabular}{|c|c|c|c|}
    \hline
    \textbf{Datasets} & \textbf{$v$} & \textbf{$n_\mu$} & \textbf{$\mu$} \\
    \hline
    FR\_SURV \cite{rudrani2011face} & $\{0,...,7\}$ & $8$ & $\{0,...,7\}$ \\
    \hline
    SCFace \cite{grgic2011scface} & $\{0,...,5\}$ & $16$ & $\{0,...,15\}$ \\
    \hline
    ChokePoint \cite{wong_cvprw_2011} & $\{0,...,4\}$ & $4$ & $\{0,...,3\}$ \\
    \hline
\end{tabular}
\end{center}
\label{tab:gab}
\end{table}

\subsubsection{Weber Faces}
\label{wf}

Weber’s law suggests that for a stimulus, the ratio between the smallest perceptual change and the background is a constant, which implies stimuli are perceived not in absolute terms but in relative terms. Inspired from this,  a novel illumination insensitive representation of face images is exploited and analyzed under varying illuminations via a ratio image, called “Weber-face,” \cite{wang2011illumination} where a ratio between local intensity variation and the background is computed.

\subsubsection{Local Binary Pattern}
\label{lbp}

Local binary patterns (LBP) \cite{ahonen2006face} is a type of feature used for classification in computer vision. LBP is the particular case of the Texture Spectrum model proposed in 1990. The face image is divided into several regions from which the LBP feature distributions are extracted and concatenated into an enhanced feature vector to be used as a face descriptor. The procedure consists of using the texture descriptor to build several local descriptions of the face and combining them into a global description. The operator assigns a label to every pixel of an image by thresholding the $3X3$-neighborhood of each pixel with the center pixel value and considering the result as a binary number. Then, the histogram of the labels can be used as a texture descriptor. 

\subsubsection{Bag-of-Words (BOW)}
\label{bow}

The feature extraction method proposed here is based on the BOW \cite{filliat2007visual} based on Dense-SIFT features. The dense-SIFT features are calculated with a single-pixel shift of the window over the face. The words used in processing are local image features. They may be constructed around interest points such as scale-space extrema (e.g. SIFT keypoints \cite{lowe2004distinctive}), or simply on windows extracted from the image at regular positions and various scales. The features can be image patches, histograms of gradient orientations or color histograms. As these features are sensitive to noise and are represented in high dimension spaces, they are not directly used as words, but are categorized using a vector quantization technique such as k-means. The output of this discretization is the dictionary.

\subsubsection{Fisher Vector Encoding on dense-SIFT features (FV-SIFT)}
\label{fvs}

This encoding \cite{perronnin2010improving} serves a similar purposes: summarizing in a vectorial statistic a number of local feature descriptors (e.g. SIFT  \cite{lowe2004distinctive}). Similarly to bag of visual words, they assign local descriptor to elements in a visual dictionary, obtained with a Gaussian Mixture Models for Fisher Vectors. However, rather than storing visual word occurrences only, the representation stores a statistics of the difference between dictionary elements and pooled local features. The Fisher encoding uses GMM to construct a visual word dictionary.

\subsubsection{VLAD encoding on dense-SIFT features (VLAD-SIFT)}
\label{vls}

The Vector of Linearly Agregated Descriptors \cite{arandjelovic2013all} is similar to Fisher vectors, but (i) it does not store second-order information about the features and (ii) it typically use KMeans instead of GMMs to generate the feature vocabulary (although the latter is also an option). VLAD is constructed as follows: regions are extracted from an image using an affine invariant detector, and described using the $12$8 dimensional SIFT descriptor. Each descriptor is then assigned to the closest cluster of a vocabulary of size $k$ (where, $k$ is typically $64$ or $256$, so that clusters are quite coarse). For each of the $k$ clusters, the residuals (vector differences between descriptors and cluster centers) are accumulated, and the $k$ - $128$ dimensional sums of residuals are concatenated into a single $k \times 128$ dimensional descriptor.

\subsection{Kernel Selection by Multiple Kernel Learning}
\label{ks}

\begin{figure}
\centering
\includegraphics[scale = 0.3]{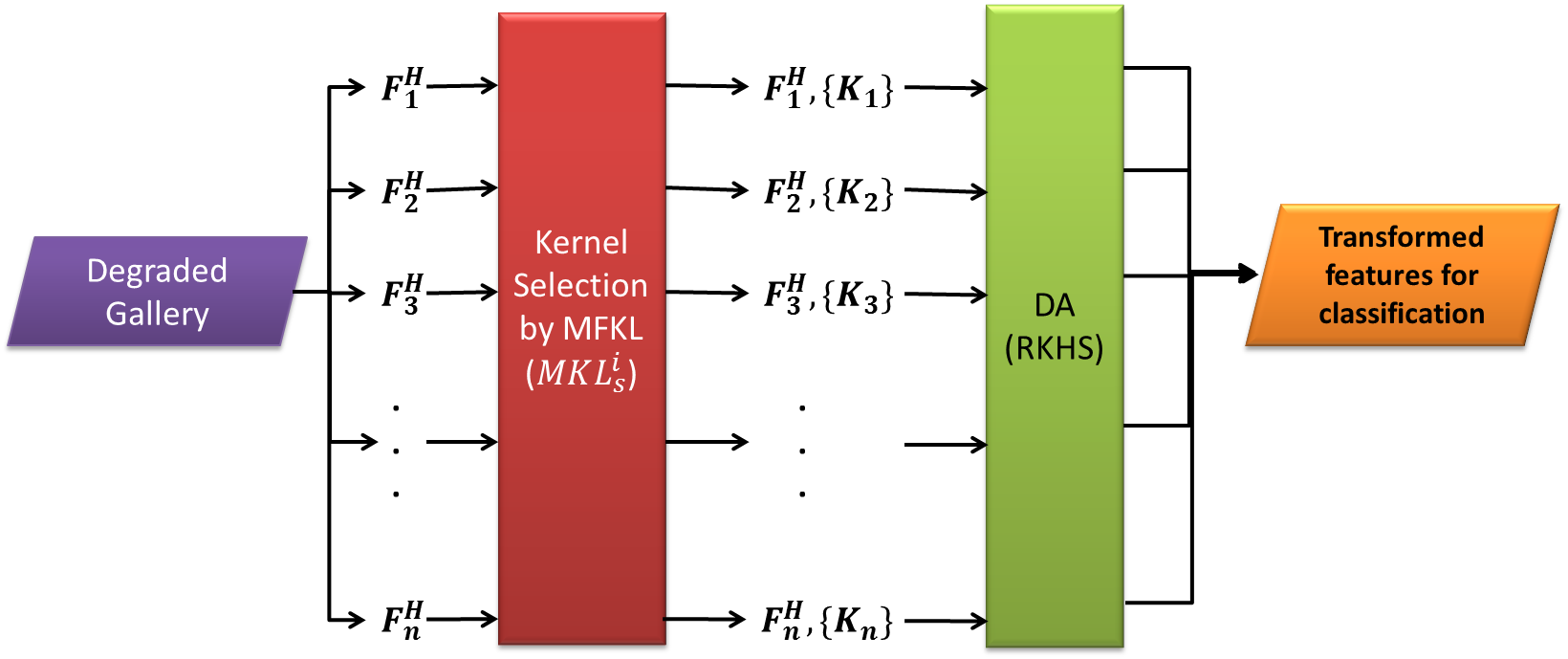}
\caption{The Training Phase after pre-processing (see figure \ref{pre_proc}), to generate transformed features in RKHS for classification.}
\label{tr}
\end{figure}

In support vector machine (SVM), selecting the kernel function and its parameters are important issues during training. Generally, to select the best performing kernel among the set of kernel function (like Linear, RBF, etc.), a cross validation procedure is used. In recent years, several MKL techniques have been proposed, where instead of selecting one specific kernel function and its corresponding parameters, multiple kernels are learned. MKL has two main advantages: (a) Different kernels correspond to different notions of similarity and instead of finding which works best, a MKL learning method helps to pick the best kernel or a combination of kernels; and (b) Different kernels may use inputs coming from different representations, possibly from different sources. In such cases, combining kernels is one possible way to combine sources of multiple information. The training phase (see figure \ref{tr}) is described in section \ref{tr_ph}.

In this paper, we introduce a novel technique based on MKL, termed MFKL (Multi-feature Kernel Learning), for selecting the optimal feature-kernel combination for classification. The MFKL method determines the optimal weights for the different kernels used for each feature category. All the features are extracted individually from each of the gallery face images and passed into the MFKL module for optimal kernel selection for each feature.  The ordered pair of the feature and kernel $<F_H^i, K_i>$ is selected and stored for further processing. We assume that we have $p$ features and their corresponding feature space be represented by $\mathcal{X}^f$, where $f \in \{1,..,p\}$. Data points $(x_i^f,y_i)$ are given, where $x_i^f \in \mathcal{X}^f$ represents a feature vector in a particular feature space $\mathcal{X}^f$, and $y_i \in \{-1, 1\}$, $\forall i \in {1,...,n} $, are the class-labels. For each feature space $\mathcal{X}^f$, the choice is one out of $m$ kernels $K_j^f \in \mathbf{R}^{n \times n}$. 

We consider $\mathcal{X} = \bigcup_{f=1}^p \mathcal{X}^f$ and $\mathcal{X} \in \mathbf{R}^l$, where $l = l_1+...+l_p$, such that $x_i^f \in \mathbf{R}^{l_f}$, where $l_f$ is the dimensionality of the feature vector $f$. This problem follows a similar formulation as described in the classification algorithm, "support kernel machine", as proposed by Bach et al. \cite{bach2004multiple}, and also discussed in section \ref{sec:mfkl}.

The primal problem is given by equation \ref{eq:mfkl_p} and its dual by equation \ref{eq:mkl_d}, with the same KKT conditions as mentioned in section \ref{sec:mfkl}. In RKHS, we assume the embeddings of the data points $x_i$ in each feature space via a mapping $\phi : \mathcal{X}^f \rightarrow \mathbf{R}^c$. We also assume that $\phi(x)$ has $m$ distinct block components $\phi(x) = (\phi_1(x), ... , \phi_m(x))$. Following the usual recipe for kernel methods, we assume that this embedding is performed implicitly, by specifying the inner product in $\mathbf{R}^c$ using a kernel function, which in this case is the sum of individual kernel functions on each block (subspace) component:

\begin{equation}
\label{eq:mfkl_k1}
\begin{split}
k_f(x_i^f,x_j^f) = \phi(x_i^f)^T \phi(x_j^f) & = \sum_{s=1}^{m} \phi_s(x_i^f)^T \phi_s(x_j^f) \\
&= \sum_{s=1}^{m} k_s(x_i^f,x_j^f)
\end{split}
\end{equation}

Now, in the feature space, $\mathcal{X}$, we have 
\begin{equation}
\label{eq:mfkl_kf1}
k(x_i,x_j) = \sum_{f=1}^p \beta_f k_f(x_i^f,x_j^f)
\end{equation}


We now “kernelize” the problem described in equation \ref{eq:mfkl_p} using this kernel function. In particular, we consider the dual of the problem in equation \ref{eq:mfkl_p} and substitute the kernel function for the inner products in the equation \ref{eq:mkl_d} with the constraint in a particular feature space, rather than over the whole space, as $(\alpha^T D(y) K_j^f D(y) \alpha) ^ {\frac{1}{2}}  \leq d_j \gamma, \forall j,f$, where $K_j^f$ is the $j$-th Gram matrix of the points $\{x_i^f\}$ corresponding to the $j$-th kernel, for the $f$-th feature formed using $k(x_i, x_j)$. These constraints are derived from equations \ref{eq:mkl_d} and \ref{eq:mkl_dk}, which lead to the simultaneous selection of feature and its corresponding non-zero kernels, based on the objective function (similar to equation \ref{eq:mkl_s}), formulated as:
\begin{equation}
\label{eq:mfkl_s}
\begin{split}
\text{min} \hspace{4pt} & \text{max}_j \frac{1}{2d^2_j} \alpha^T D(y) K_j^f D(y) \alpha - \alpha^T e \\
\text{w.r.t.} \hspace{4pt} & \alpha \in \mathbf{R}^n \\
\text{s.t.} \hspace{4pt} & 0 \leq \alpha \leq C, \alpha^T y=0 \\
\end{split}
\end{equation}
Since the sparsity of the weights of the kernels is ensured by KKT conditions, the non-zero kernels are used for classification in the testing phase. Let $J_j^f(\alpha)$ denote $\frac{1}{2d^2_j} \alpha^T D(y) K_j^f D(y) \alpha - \alpha^T e$ (see equation \ref{eq:mfkl_s}) and $J^f(\alpha) = \text{max}_j J_j^f(\alpha)$. Minimization of $J_f(\alpha)$ now reduces to an convex optimization problem, as $J_f(\alpha)$ is also a non-differentiable convex function subject to linear constraints. Our global objective function is $J(\alpha) = \bigcup_{f=1}^p J_f(\alpha)$. Union of convex functions is not necessarily convex. Hence, a subgradient method \cite{kim1991convergence} is used to solve each of these convex optimization sub-problems, $J_f(\alpha)$, and finally the union of these are used to obtain a global solution. Sparse solutions ensure that most of the kernel weights are negligible (go near to zero) and a very few non-zero kernel weights remain for each feature.

In the proposed work, we take into account a set of kernel functions for the MKL method. The set of kernels consists of Linear, Polynomial, Gaussian, RBF, Chi-square and RBF + Chi-square. The equations of each of these kernels are tabulated in Table \ref{tab:ker_typ}. 

\begin{table}[!htbp]
\caption{Different types of kernel used in the MFKL, with their formulae.}
\begin{center}
\begin{tabular}{|c|c|}
    \hline
    \textbf{Type of Kernel} & \textbf{Formula} \\
    \hline
    Linear & $k(x,y) = x^T y + c$ \\
    \hline
    Polynomial & $k(x,y) = (\alpha x^T y + c) ^d$ \\
    \hline
    Gaussian & $k(x, y) = exp \left(-\frac{\| x-y \| ^2}{2\sigma^2} \right)$ \\
    \hline
    RBF & $k(x, y) = exp \left(-\frac{\| x-y \|}{2\sigma^2} \right)$ \\
    \hline
    Chi-square & $k(x,y) = 1 - \Sigma_{i=1}^n \frac{(x_i-y_i)^2}{\frac{1}{2}(x_i+y_i)}$ \\
    \hline
    RBF + Chi-square & $k(x, y) = 1 - \Sigma_{i=1}^n \frac{(x_i-y_i)^2}{\frac{1}{2}(x_i+y_i)}$\\ 
    & $+ exp \left(-\frac{\| x-y \|}{2\sigma^2} \right)$ \\
    \hline
\end{tabular}
\end{center}
\label{tab:ker_typ}
\end{table}

\subsection{The Training Phase}
\label{tr_ph}
In this phase of our proposed framework as shown in figure \ref{tr}, we have a set of feature $\mathcal{F}$ and a set of kernels $\mathcal{K}$ pairings.
\begin{equation}
\begin{split}
\mathcal{F}= & \{LBP, Eigen Faces, Fisher Faces, Gabor faces, \\
& Weber Faces, VLAD-SIFT, FV-SIFT, BOW\}
\end{split}
\end{equation}
where each $F_i \in \mathcal{F}$ is the feature extracted from a face image; and
\begin{equation}
\begin{split}
\mathcal{K}= & \{Linear, Polynomial, Gaussian, RBF, \\ 
& Chi-square, RBF + Chi-square\}
\end{split}
\end{equation}
where each $K_i \in \mathcal{K}$ is a kernel function for the projection in the RKHS, $\mathcal{H}_i$.

The combination of $\mathcal{F}$ and $\mathcal{K}$ is passed into the MKL module to obtain the set of optimized pair of $\{F_j, K_j\}$ using the kernel selection method described in section \ref{ks}. Based on the best feature-kernel pair obtained, the feature vector is projected into a higher dimensional space of RKHS. The training in DA is performed to obtain the final model parameters along with the feature-kernel pairs. The number of probe samples used as targets for DA is mentioned in the table \ref{tab:sigda}. 

\subsection{Classification based on K-Nearest Neighbor Classifier}

\begin{figure}[!htbp]
\centering
\includegraphics[scale = 0.35]{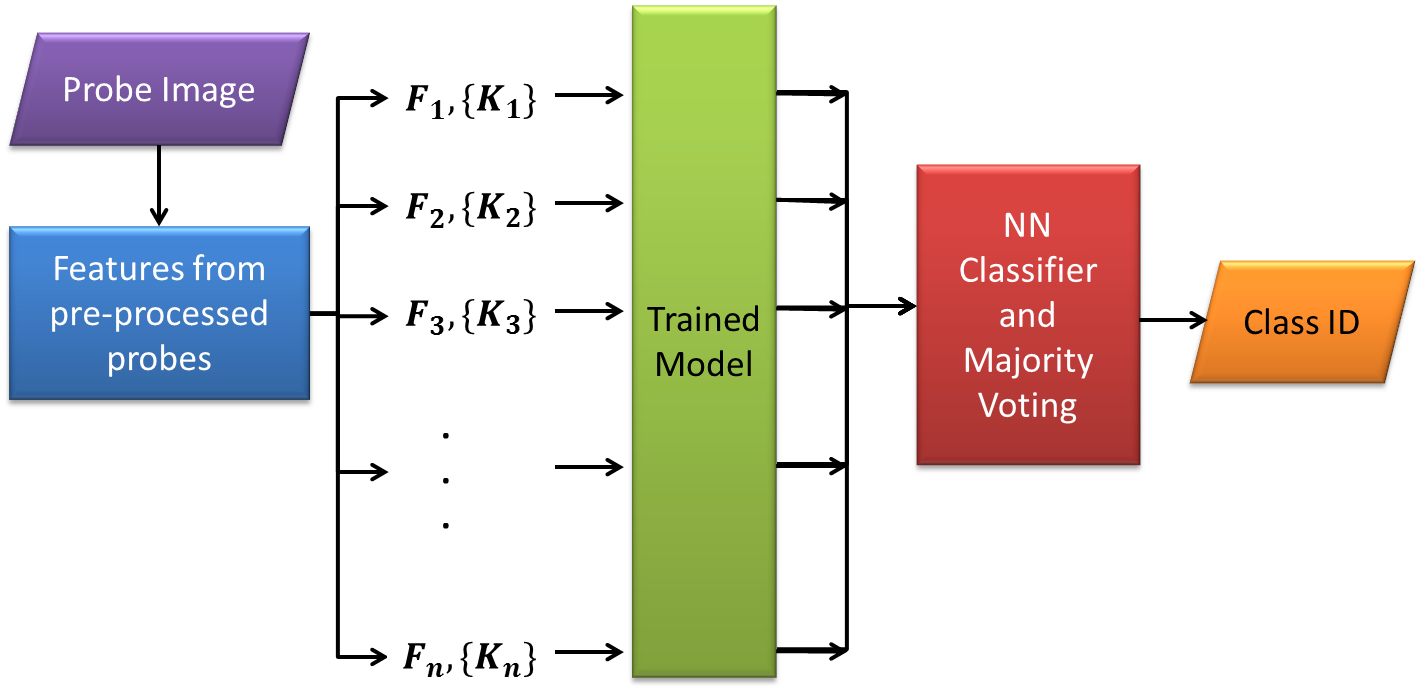}
\caption{The Testing phase in the proposed method (see figure \ref{pffig}).}
\label{tes_ph}
\end{figure}

In the testing phase as shown in figure \ref{tes_ph}, a query low-resolution face image is first pre-processed based on the pre-processing techniques described in section \ref{pps}. Features are extracted from the pre-processed probe and passed into the DA module (RKHS), as shown in figure \ref{tes_ph}. The process of kernel selection corresponding to a feature is based on the MFKL technique proposed in the kernel selection stage. The overall gram matrix is created for each of these features. A majority voting is used based on the Nearest neighbor classification for the probe samples. The winning class ID is selected as the best match. 

\section{Details of Surveillance Face Databases Used}
For the experimentation purpose we have used three real-world surveillance face datasets, which are discussed below. In all three real-world datasets the gallery samples are taken in laboratory conditions, while for probes two out of the three datasets are shot indoor, while one is shot outdoor.
\subsection{FR\_SURV \cite{rudrani2011face}}
\label{frs}

\begin{figure}[!htbp]
\centering
\includegraphics[scale=0.5]{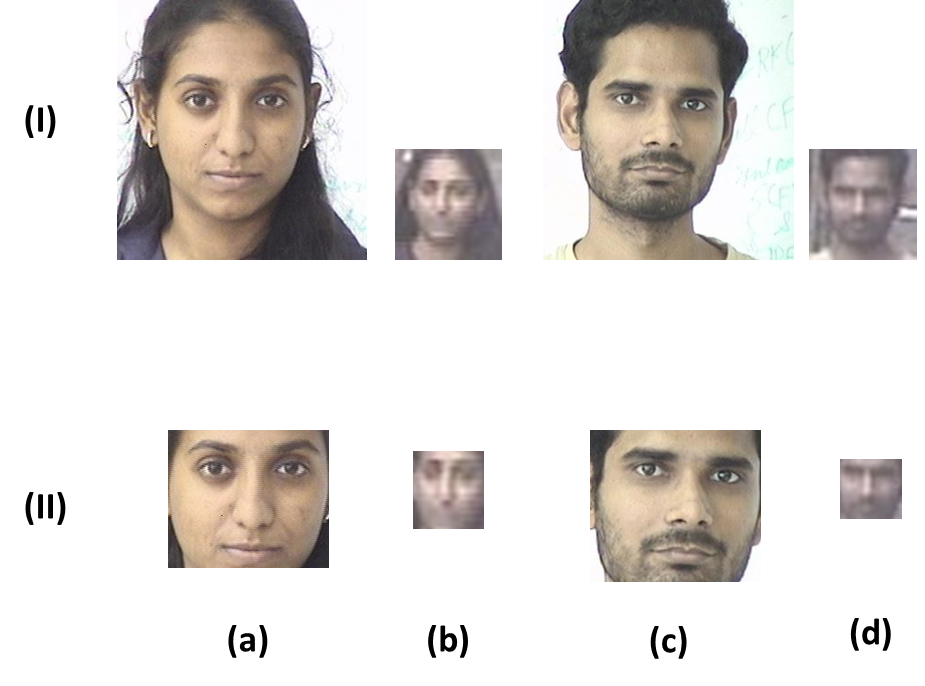}
\caption{Samples of two subjects from FR\_SURV Database \cite{rudrani2011face}: (I) (a), (c) Gallery images; and (I) (b), (d) corresponding Probe images; (II) (a)-(d) Cropped faces from (I) using Chehra \cite{asthanaincremental}.}
\label{itsfig1}
\end{figure}

FR\_SURV is a challenging database for FR, because the gallery and the probe images are taken at different resolutions with two different cameras. The gallery samples, taken indoor with high resolution camera, have a resolution of $250 \times 250$ pixels, while the probe samples, taken by surveillance camera, have a very low resolution of $45 \times 45$ pixels. The probe samples are taken at a distance of 50-100 meters outdoor. Using \textit{Chehra} \cite{asthanaincremental} on both the gallery and probe samples, we produce cropped face regions at an average of $150 X 150$ pixels and $33 \times 33$ pixels respectively. The database consists $51$ subjects with $20$ samples per class. Figure \ref{itsfig1} shows two samples of the gallery images and the their respective probe image (cropped using \textit{Chehra} \cite{asthanaincremental}).

\subsection{SCFace \cite{grgic2011scface}}
\label{scf}

\begin{figure}[!htbp]
\centering
\includegraphics[scale=0.3]{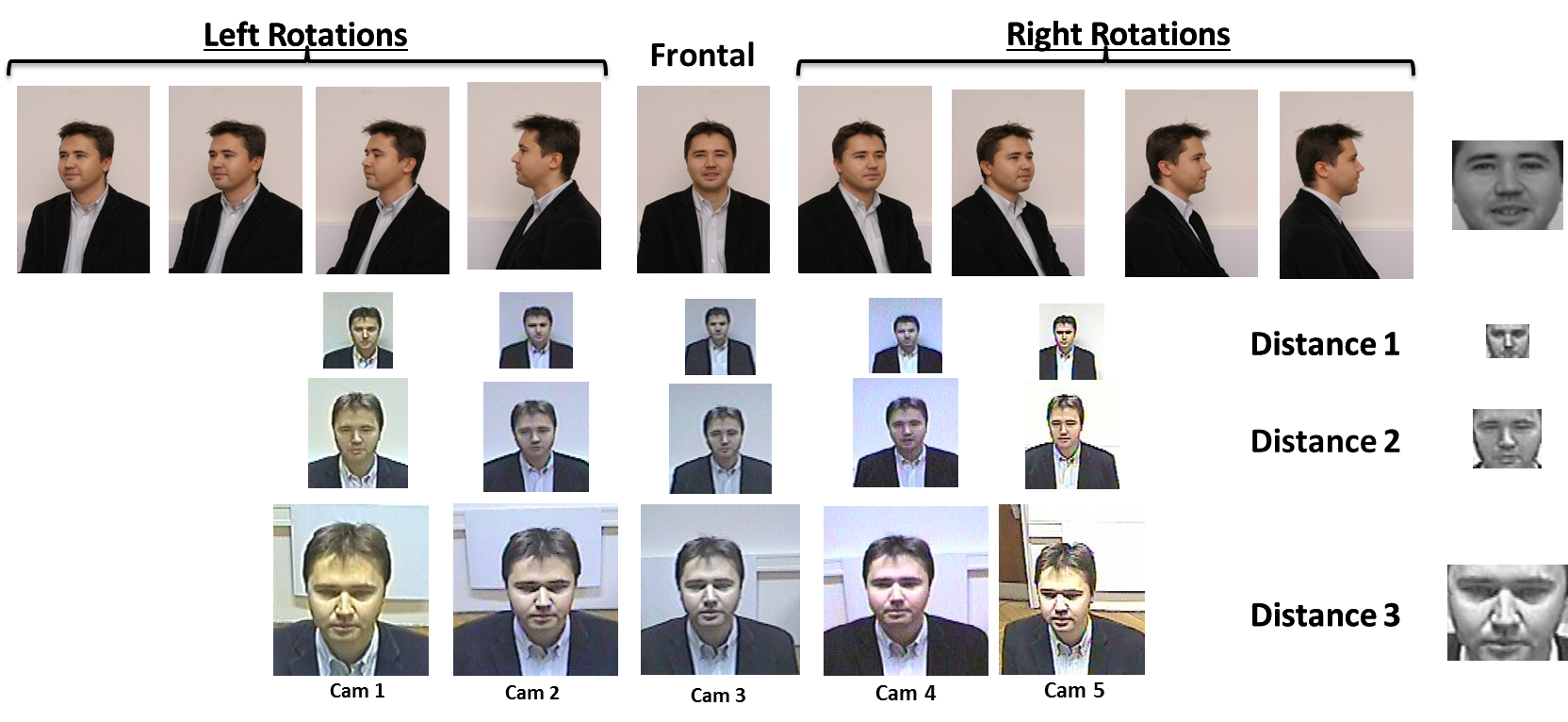}
\caption{SCface Database \cite{grgic2011scface}: (I) Gallery images; Probe Images at (II) \textit{Distance 1}; (III) \textit{Distance 2}; and (IV) \textit{Distance 3}; Right column shows the \textit{Chehra} \cite{asthanaincremental} output for any one of the samples in each row.}
\label{scffig}
\end{figure}

SCface is also a challenging database for FR as the images were taken at different surveillance conditions. The database has a huge collection of static images of $130$ different people. Images were captured by five different video surveillance cameras (cam1, cam2, cam3, cam4, cam5). Two cameras were also used in the night vision mode (cam6 and cam7). All these images were collected indoor with varying quality and resolution levels at three different distances. The training set consists of nine images: one frontal and four each in left and right rotations. The dataset has an image taken from an infra-red camera(cam8). Figure \ref{scffig} shows the images for a single person in the dataset. The gallery has images of size $2048 \times 3072$ pixels which are cropped by VJFD to an average of $800 \times 600$ pixels. The probe images at \textit{Distance 1,2} and \textit{3} has a resolution of $75 \times 100, 108 \times 144$ and $168 \times 224$ pixels respectively which are cropped by \textit{Chehra} \cite{asthanaincremental} to an average of $40 \times 40, 60 \times 60$ and $100 \times 100$ pixels respectively. We do not use the cam6 and cam7 as they are IR images.  

\subsection{ChokePoint \cite{wong_cvprw_2011}}
\label{cp}

\begin{figure} [!htbp]
\centering
\includegraphics[scale=0.75]{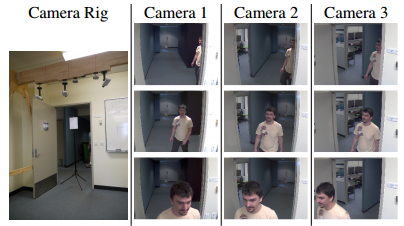}
\caption{An example of the recording setup used for the ChokePoint dataset \cite{wong_cvprw_2011}. A camera rig contains 3 cameras placed just above a door, used for simultaneously recording the entry of a person from 3 viewpoints. The variations between viewpoints allow for variations in walking directions, facilitating the capture of a near-frontal face by one of the cameras.}
\label{cp_ex1}
\end{figure}

\begin{figure} [!htbp]
\centering
\includegraphics[scale=0.6]{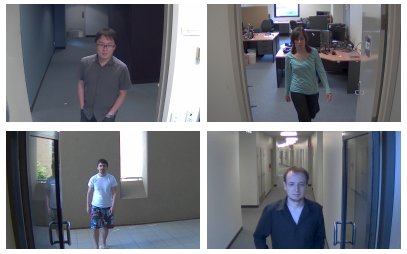}
\caption{Example shots from the ChokePoint dataset \cite{wong_cvprw_2011}, showing portals with various backgrounds.}
\label{cp_ex2}
\end{figure}

Wong et al. \cite{wong_cvprw_2011} collected a new video dataset, termed ChokePoint, designed for experiments in person identification/verification under real-world surveillance conditions using existing technologies. An array of three cameras was placed above several portals (natural choke points in terms of pedestrian traffic) to capture subjects walking through each portal in a natural way (see figures \ref{cp_ex1} and \ref{cp_ex2}).

The dataset consists of 25 subjects (19 male and 6 female) in portal 1 and 29 subjects (23 male and 6 female) in portal 2. In total, it consists of 48 video sequences and 64,204 face images. Each sequence was named according to the recording conditions (eg. P2E S1 C3) where P, S, and C stand for portal, sequence and camera, respectively. E and L indicate subjects either entering or leaving the portal. The numbers indicate the respective portal, sequence and camera label. For example, P2L S1 C3 indicates that the recording was done in Portal 2, with people leaving the portal, and captured by camera 3 in the first recorded sequence. The ChokePoint dataset does not have variation in resolution. But the difference lies in the different cameras used for capturing the image due to different camera parameters and the illumination  variations. This feature makes it suitable to be tackled with DA. 

For experimentation, we consider the images obtained from camera, C1 as the Gallery set, since it contains maximum frontal images with better lighting conditions. The other cameras are considered as probe images. We do not consider the sequence S5, as it contains crowded scenario. Since the images obtained from C1 are crisp and have better illumination conditions than C2 and C3, the gallery and probe is passed through all the pre-processing stages, except the face hallucination stage, as the resolution of the images obtained from all these cameras are similar. 

\subsection{Intermediate results of face processing}

\begin{figure} [!htbp]
\centering
\includegraphics[scale=0.35]{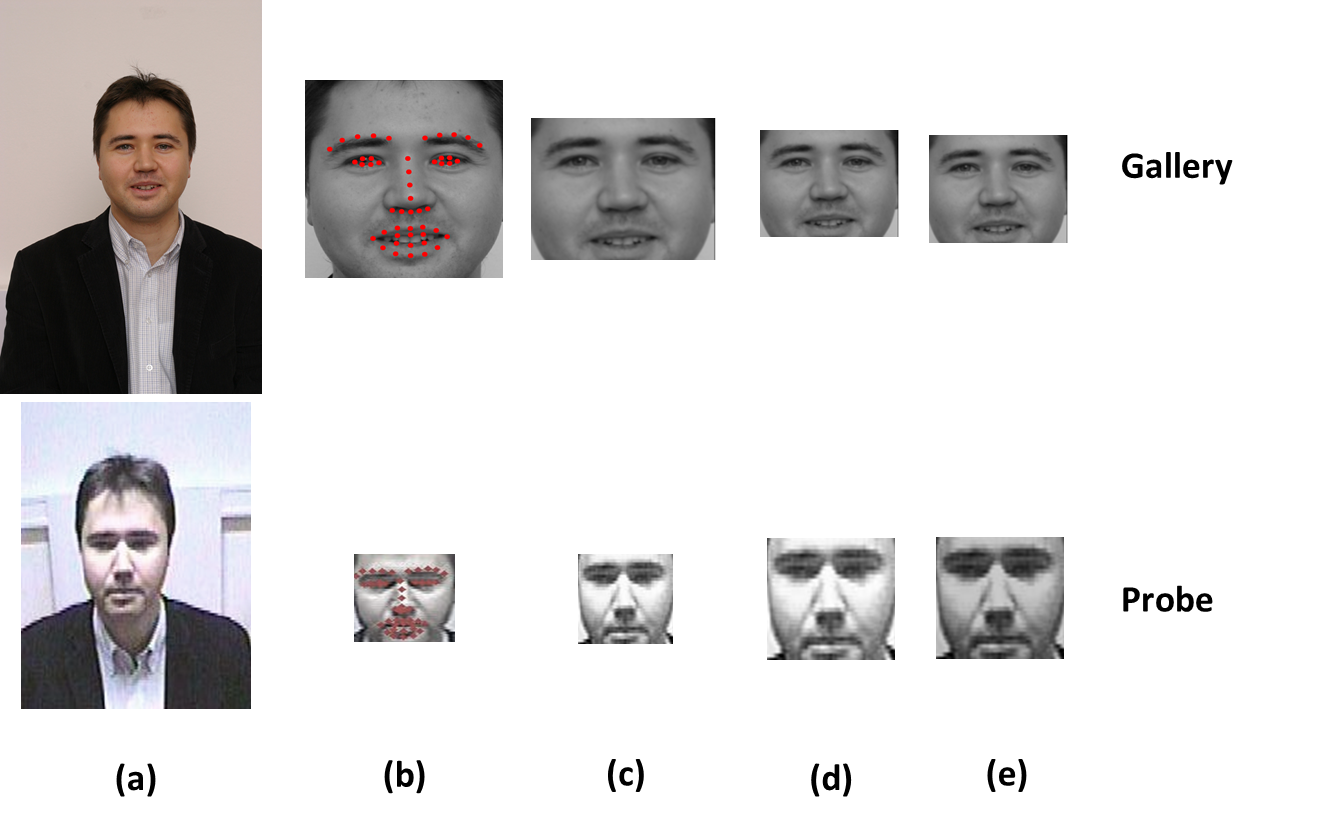}
\caption{Pre-processing stages on SCFace \cite{grgic2011scface} for a subject: (a) Original image, (b) Landmarks detected by Chehra \cite{asthanaincremental}, (c) the cropped faces obtained using Chehra, (d) Downsampled image of gallery and upsampled image of probe, (e) Final degraded gallery and enhanced probe images. }
\label{sc_int}
\end{figure}

The face images go through the several stages of pre-processing, as described in section \ref{pps}. An example to illustrate the pre-processing stages is shown in figure \ref{sc_int}, for the SCFace \cite{grgic2011scface} dataset. The top row illustrates the effect of pre-processing on the gallery images, while the bottom row illustrates the effect of pre-processing on the probe images. Figure \ref{sc_int}(a) shows the original images available in the dataset, while figure \ref{sc_int}(b) shows the landmark points detected by the \textit{Chehra} \cite{asthanaincremental} on the face images. Based on these landmark points, we obtain a tightly cropped face image as shown in figure \ref{sc_int}(c). Figure \ref{sc_int}(d) shows the result of downsampling of the gallery images and upsampling of the probe images by Face Hallucination technique, as discussed in section \ref{fh}. The downsampled gallery images are blurred using a Gaussian kernel to obtain the degraded gallery images while the upsampled probe images are passed through Power law transformation to obtain moderately enhanced probe images as shown in figure \ref{sc_int}(e). Face hallucination is not applied on the probe samples in the ChokePoint \cite{wong_cvprw_2011} dataset, as the resolution of the gallery and probe samples are similar in the dataset. 

%

\begin{figure} [!htbp]
\centering
\includegraphics[scale=0.3]{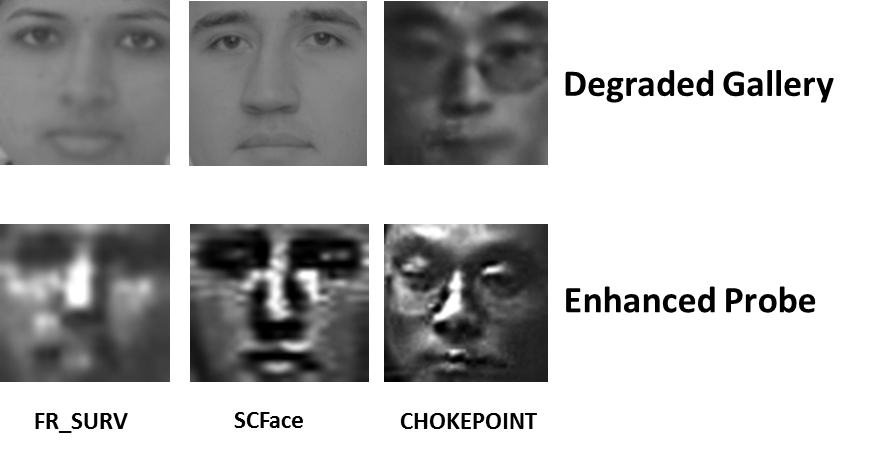}
\caption{Example shots from the three datasets (one sample each), showing the degraded gallery and the enhanced probe images on the cropped faces produced by Chehra \cite{asthanaincremental}}
\label{gp}
\end{figure}

Figure \ref{gp} shows the examples of the degraded gallery (in the top row) and the enhanced probe (in the bottom row) images of a single subject from the three surveillance datasets, used in our experimentation. These degraded gallery and the enhanced probe images are used for feature extraction. The original gallery and probes are also shown with similar spatial resolution to illustrate the efficiency of the pre-processing stage (ignore resolution which is different).

\section{Experimental Results}

Rigorous experimentation is carried on three real-world datasets; \textit{SCface}\cite{grgic2011scface}, \textit{FR\_SURV}\cite{rudrani2011face}, and \textit{ChokePoint}\cite{wong_cvprw_2011}  The proposed methods are compared with several other recent state-of-the-art methods and the results are reported in table \ref{tab:res} using Rank-1 Recognition Rate.

\begin{table}[!htbp]
\caption{Rank-1 Recognition Rate for different methods. Results in bold, exhibit the best performance.}
\begin{tabular}{@{}|p{.3cm}|p{2cm}|p{1cm}|p{1.5cm}|p{1.7cm}|} 
  \hline
  \textbf{Sl.} & \textbf{Algorithm} & \textbf{SCface} & \textbf{FR\_SURV} & \textbf{ChokePoint}  \\
  & & \cite{grgic2011scface} & \cite{rudrani2011face} &  \cite{wong_cvprw_2011} \\\hline
  1 & EDA1 \cite{banerjee2014face} & 47.65 & 7.82 & 54.21 \\\hline
  2 & COMP\_DEG \cite{rudrani2011face}& 4.32 & 43.14 & 62.59 \\\hline
  3 & MDS \cite{biswas2012multidimensional}& 42.26 & 12.06 & 52.13 \\\hline
  4 & KDA1 \cite{banerjee2014face} & 35.04 & 38.24 & 56.25 \\\hline
  5 & Gopalan \cite{gopalan} & 2.06 & 2.06 & 58.62 \\\hline
  6 & Kliep \cite{kliep} & 37.51 & 28.79 & 63.28 \\\hline
  7 & Naive & 75.27 & 45.78 & 65.76 \\\hline
  8 & Proposed Method & \textbf{78.31} & \textbf{55.23} & \textbf{84.62} \\\hline
  
\end{tabular}
\label{tab:res}
\end{table}

In case of Naive combination, the source and the target domain samples are used without transformation for training and domain samples are used as probes. In EDA1 method, proposed by Banerjee et al. \cite{banerjee2014face}, DA processing based on an eigenvector based transformation, whose extension in the RKHS is termed as KDA1.  Rudrani et al. \cite{rudrani2011face} (COMP\_DEG) tries to bridge the gap between the gallery and the probe samples by projecting them both into a lower dimensional subspace determined by the principal components of the feature vectors obtained from each face. This paper also acts as the source for the dataset, FR\_SURV \cite{rudrani2011face}. Multi-dimensional scaling (MDS) proposed by Biswas et al. in \cite{biswas2012multidimensional} projects both the gallery and the probe samples into a common subspace for classification. The methods proposed by Gopalan et al.\cite{gopalan} and Kliep \cite{kliep} are two DA based techniques used for object classification accross domains. We can observe that the method proposed in this paper have outperformed (our results are given in bold, in Table \ref{tab:res}) all the other competing methods by a considerable margin. The complexity of the datasets is also observed by the results of FR, which are all moderately low in many cases.

\begin{figure} [!htbp]
\centering
\includegraphics[scale=0.3]{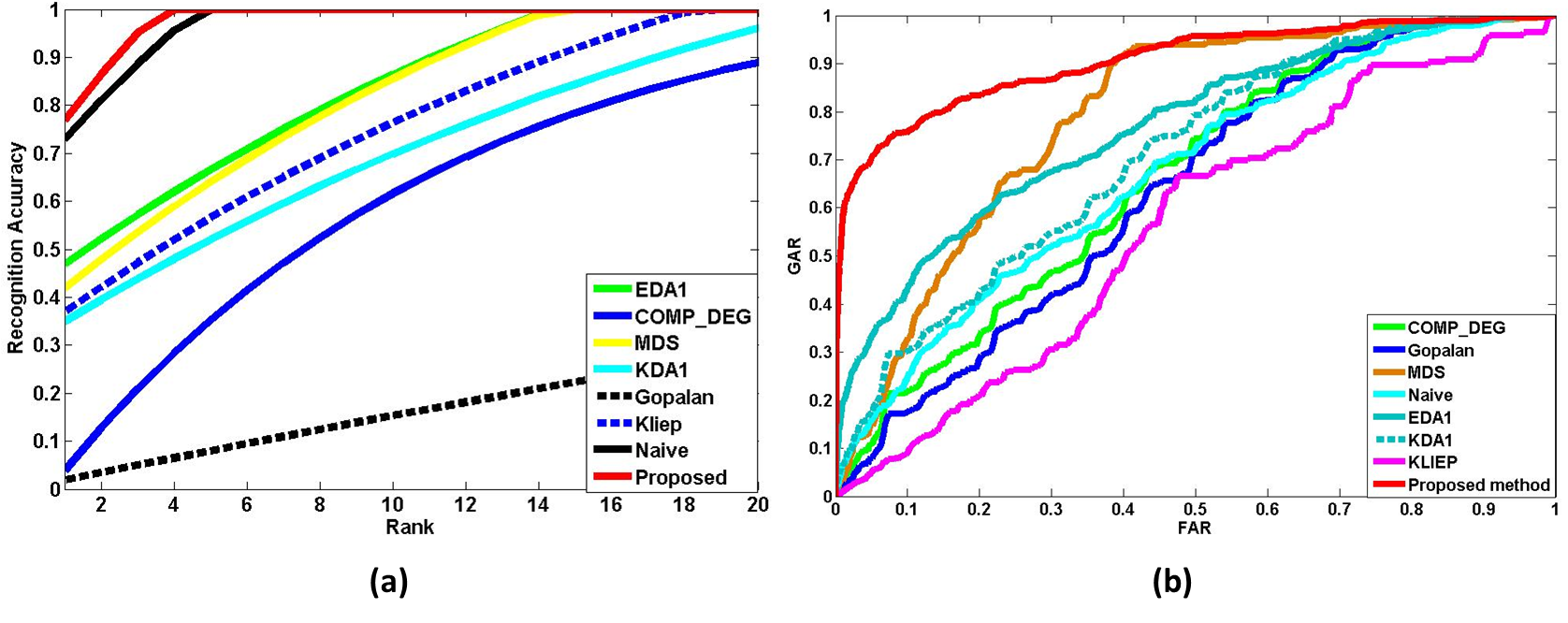}
\caption{(a) ROC and (b) CMC plots for performance analysis of different methods, using SCFace \cite{grgic2011scface} dataset.}
\label{exp_scf}
\end{figure}

\begin{figure} [!htbp]
\centering
\includegraphics[scale=0.32]{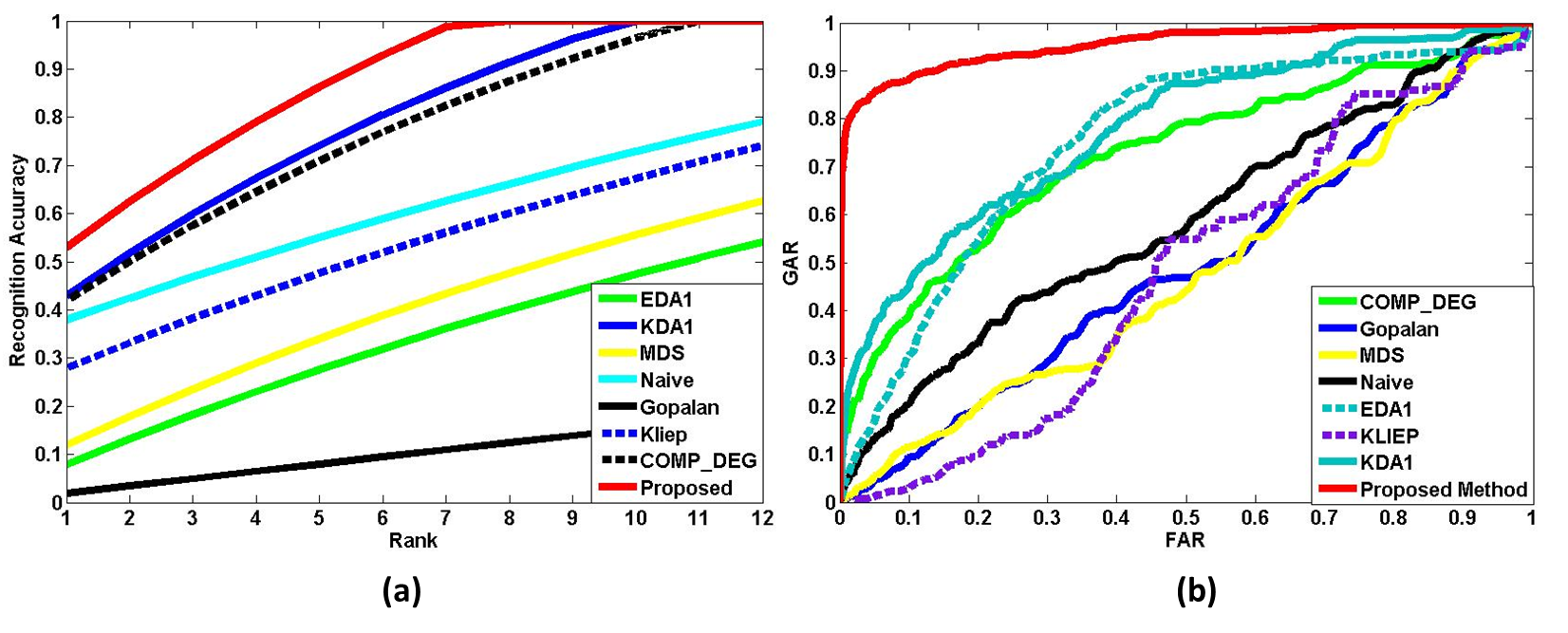}
\caption{(a) ROC and (b) CMC plots for performance analysis of different methods, using FR\_SURV \cite{rudrani2011face} dataset.}
\label{exp_frs}
\end{figure}

\begin{figure} [!htbp]
\centering
\includegraphics[scale=0.34]{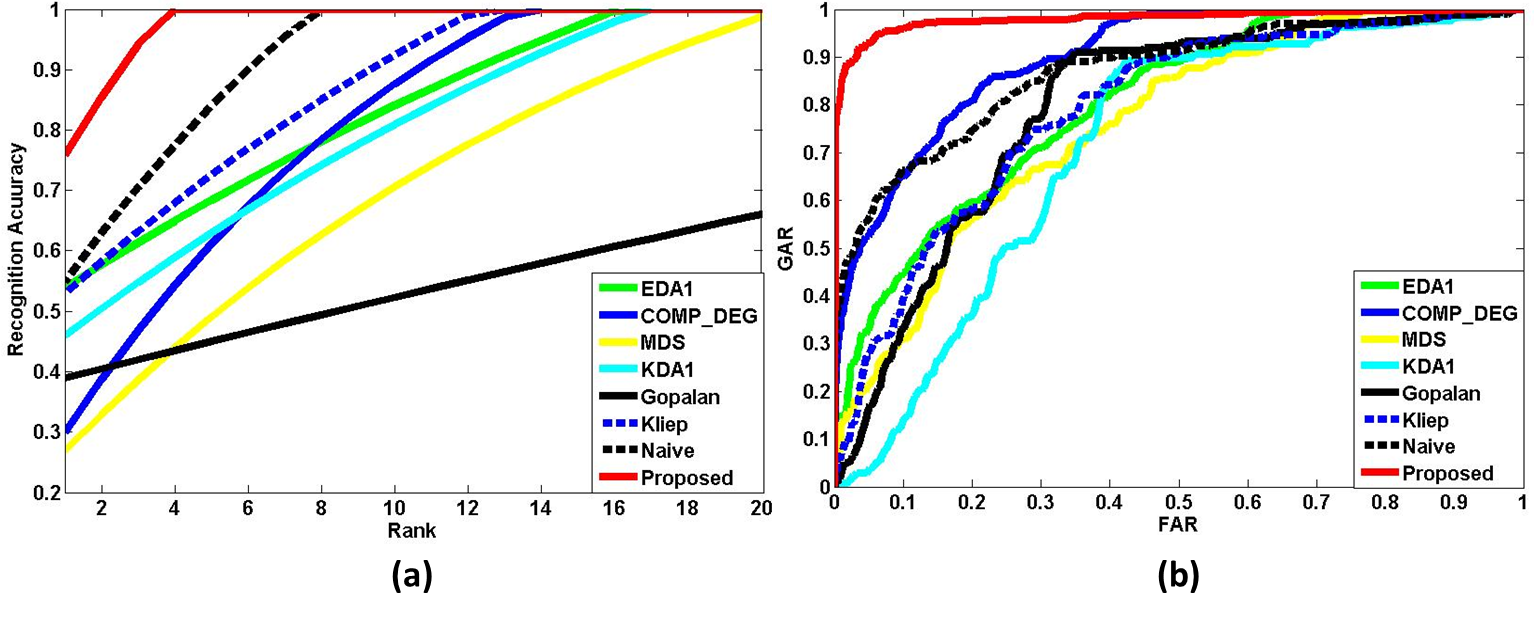}
\caption{(a) ROC and (b) CMC plots for performance analysis of different methods, using ChokePoint \cite{wong_cvprw_2011} dataset.}
\label{exp_cp}
\end{figure}

Results are also reported using ROC (for identification) and CMC (for verification) measures, as shown in figures \ref{exp_scf} - \ref{exp_cp}, for the three datasets respectively. The plot drawn in red show the performance of our proposed method. We can observe that the red curves in all the plots outperform all other competing methods. On an average, the second best performance is given by the naive approach since the MFKL is also incorporated into it, while the method proposed by Gopalan et al. \cite{gopalan} performs generally the worst.

As we look closer into the the table \ref{tab:res} row-wise, we can easily observe that the FR\_SURV datasets has the least accuracy. This is an indication that there is still further scope of improvement in this field. Also, it shows that the database is quite tough to handle. As we can see that the gallery samples in FR\_SURV are all taken in Indoor laboratory conditions and the probe samples are taken in Outdoor conditions which results in the large complexity of the database. Since FR\_SURV is an outdoor dataset, we can see the accuracy of FR is less than that of the ChokePoint dataset, which is the easiest to handle among the three. The SCface and the ChokePoint datasets are two indoor surveillance datasets. Experiments are done in both identification and verification mode. There is still scope of improvement to find a more effective effective transformation such that the distribution of the features of the gallery and the probe become similar. The other two datasets  have both the gallery and the probe taken in Indoor scenario. The effectiveness of the DA used in this paper is clearly visible in the results of EDA1 and KDA1. The non-linear transformations in KDA1 proves to be more effective which motivates this paper to concentrate mostly on the DA in RKHS. The effectiveness of the MKL based method is evident when we try to focus on the Naive combination results. It is very competitive in all the three datasets. The Naive combination is the complete Framework without the DA module, incorporating also the MKL process. When these two powerful tools are combined, our proposed method outperforms all other methods by a considerable margin.

\section{Conclusion}
An efficient method to tackle the problem of low-contrast and low-resolution in face recognition under surveillance scenario is proposed in this paper. The method proposes a novel kernel selection method using MFKL to obtain an optimal pairing of feature and kernel for eigen-domain transformation based unsupervised DA in the RKHS. The three metrics used to compare the performance of our proposed method with the recent state-of-the-art techniques, show a great deal of superiority of our method than the other techniques, using three real-world surveillance face datasets.


%

\ifCLASSOPTIONcaptionsoff
  \newpage
\fi



%
%
%
\bibliographystyle{abbrv}

\bibliography{tifs_mkl}

%





\end{document}